\documentclass[10pt,twocolumn,letterpaper]{article}

\usepackage[pagenumbers]{cvpr} 

\usepackage{graphicx}
\usepackage{amsmath}
\usepackage{amssymb}
\usepackage{booktabs}
\usepackage[normalem]{ulem}

\usepackage[table]{xcolor}  
\definecolor{citecolor}{HTML}{0071bc}
\usepackage[pagebackref,breaklinks,colorlinks,citecolor=citecolor]{hyperref}

\usepackage[capitalize]{cleveref}
\crefname{section}{Sec.}{Secs.}
\Crefname{section}{Section}{Sections}
\Crefname{table}{Table}{Tables}
\crefname{table}{Tab.}{Tabs.}

\def\cvprPaperID{*****} 
\def\confName{CVPR}
\def\confYear{2023}

\usepackage{tabulary,xspace}
\usepackage{fixmath,mathtools,nicefrac,mmstyle}
\usepackage{cite}
\usepackage{booktabs}
\usepackage{subcaption}
\usepackage{caption}
\usepackage{multirow}
\usepackage[misc]{ifsym} 
\usepackage{float}

\newcommand{\authorskip}{\hspace{2.5mm}}
\renewcommand{\paragraph}[1]{\vspace{1.25mm}\noindent\textbf{#1}}
\newlength\savewidth\newcommand\shline{\noalign{\global\savewidth\arrayrulewidth
  \global\arrayrulewidth 1pt}\hline\noalign{\global\arrayrulewidth\savewidth}}
\newcommand{\tablestyle}[2]{\setlength{\tabcolsep}{#1}\renewcommand{\arraystretch}{#2}\centering\small}

\newcommand{\bd}[1]{\textbf{#1}}
\newcommand{\x}{$\times$}
\newcolumntype{x}[1]{>{\centering\arraybackslash}p{#1pt}}

\definecolor{deemph}{gray}{0.6}

\definecolor{ourscolor}{gray}{.9}
\newcommand{\ours}[1]{\cellcolor{ourscolor}{#1}}

\begin{document}

\title{RTMDet: An Empirical Study of Designing Real-Time Object Detectors}

\author{Chengqi Lyu$^{1*}$ \authorskip Wenwei Zhang$^{1,2*}$ \authorskip Haian Huang$^1$  \authorskip Yue Zhou$^{1,4}$ \authorskip Yudong Wang$^{1,3}$ \\
\authorskip Yanyi Liu$^{5}$ \authorskip Shilong Zhang$^1$ \authorskip Kai Chen$^{1}$ \\[2mm]
\small $^{*}$equal contribution \\[2mm]
$^{1}$Shanghai AI Laboratory \authorskip $^{2}$S-Lab, Nanyang Technological University \\
$^{3}$School of Electrical and Information Engineering, Tianjin University \\
$^{4}$Department of Electronic Engineering, Shanghai Jiao Tong University \\
$^{5}$Northeastern University \\
{\tt\small $\left \{\text{lvchengqi, chenkai}\right\}$@pjlab.org.cn, wenwei001@e.ntu.edu.sg } \\
}

\maketitle


\begin{abstract}

In this paper, we aim to design an efficient real-time object detector that exceeds the YOLO series and is easily extensible for many object recognition tasks such as instance segmentation and rotated object detection.
To obtain a more efficient model architecture, we explore an architecture that has compatible capacities in the backbone and neck, constructed by a basic building block that consists of large-kernel depth-wise convolutions.
We further introduce soft labels when calculating matching costs in the dynamic label assignment to improve accuracy.
Together with better training techniques, the resulting object detector, named RTMDet, achieves \bd{52.8\% AP on COCO with 300+ FPS} on an NVIDIA 3090 GPU, outperforming the current mainstream industrial detectors.
RTMDet achieves the best parameter-accuracy trade-off with tiny/small/medium/large/extra-large model sizes for various application scenarios, and obtains new state-of-the-art performance on real-time instance segmentation and rotated object detection.
We hope the experimental results can provide new insights into designing versatile real-time object detectors for many object recognition tasks.
Code and models are released at \url{https://github.com/open-mmlab/mmdetection/tree/3.x/configs/rtmdet}.
\end{abstract}

\begin{figure*}[ht]
    \centering
    \includegraphics[width=0.9\linewidth]{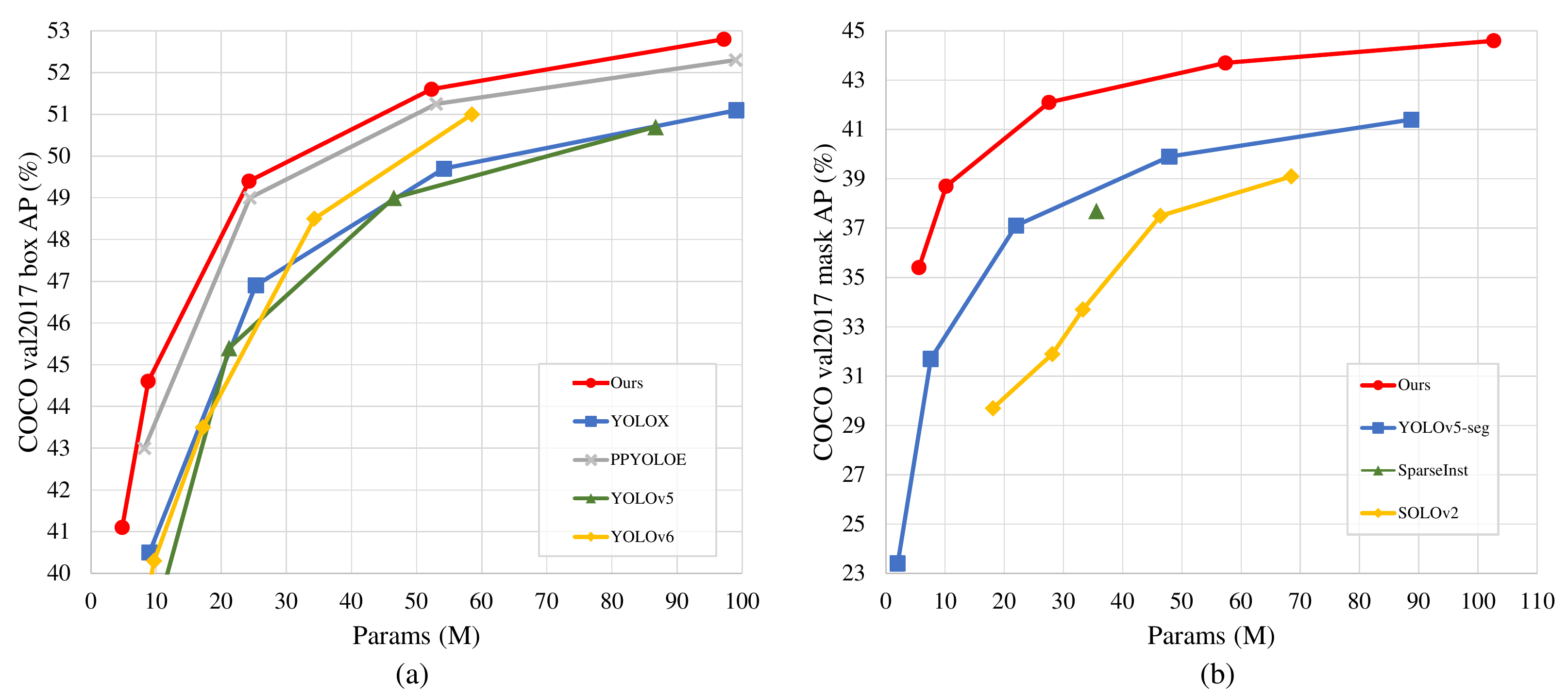}
    \vspace{-10pt}
    \caption{Comparison of parameter and accuracy. (a) Comparison of RTMDet and other state-of-the-art real-time object detectors. (b) Comparison of RTMDet-Ins and other one-stage instance segmentation methods.}
    \label{fig:compare_params}
    \vspace{-12pt}
\end{figure*}

\section{Introduction}\label{sec:Introduction}

Optimal efficiency is always the primary pursuit in object detection, especially for real-world perception in autonomous driving, robotics, and drones.
Toward this goal, YOLO series~\cite{YOLO, YOLO9000, YOLOv3, YOLOv4, YOLOv5, YOLOX, YOLOv6, YOLOv7} explore different model architectures and training techniques to improve the accuracy and efficiency of one-stage object detectors continuously.

In this report, we aim to push the limits of the YOLO series and contribute a new family of \emph{\textbf{R}eal-\textbf{T}ime \textbf{M}odels for object \textbf{Det}ection}, named \textbf{RTMDet}, which are also capable of doing instance segmentation and rotated object detection that previous works have not explored. 
The appealing improvements mainly come from better representation with large-kernel depth-wise convolutions and better optimization with soft labels in the dynamic label assignments.

Specifically, we first exploit large-kernel depth-wise convolutions in the basic building block of the backbone and neck in the model, which improves the model's capability of capturing the global context~\cite{replknet}.
Because directly placing depth-wise convolution in the building block will increase the model depth thus slowing the inference speed, we further reduce the number of building blocks to reduce the model depth and compensate for the model capacity by increasing the model width.
We also observe that putting more parameters in the neck and making its capacity compatible with the backbone could achieve a better speed-accuracy trade-off.
The overall modification of the model architectures allows the fast inference speed of RTMDet without relying on model re-parameterizations~\cite{YOLOv6, YOLOv7, PPYOLOE}.

We further revisit the training strategies to improve the model accuracy.
In addition to a better combination of data augmentations, optimization, and training schedules, we empirically find that existing dynamic label assignment strategies~\cite{YOLOX, TOOD} can be further improved by introducing soft targets instead of hard labels when matching ground truth boxes and model predictions. Such a design improves the discrimination of the cost matrix for high-quality matching but also reduces the noise of label assignment, thus improving the model accuracy.

RTMDet is generic and can be easily extended to instance segmentation and rotated object detection with few modifications.
By simply adding a kernel and a mask feature generation head~\cite{condinst, Cheng2022SparseInst}, RTMDet can perform instance segmentation with only around 10\% additional parameters.
For rotated object detection, RTMDet only needs to extend the dimension (from 4 to 5) of the box regression layer and switch to a rotated box decoder. We also observe that the pre-training on general object detection datasets~\cite{lin2014coco} is beneficial for rotated object detection in aerial scenarios~\cite{xia2018dota}.

We conduct extensive experiments to verify the effectiveness of RTMDet and scale the model size to provide tiny/small/medium/large/extra-large models for various application scenarios.
As shown in Fig.~\ref{fig:compare_params}, RTMDet achieves a better parameter-accuracy trade-off than previous methods and gains superior performance to previous models~\cite{YOLOv3, YOLOv4, YOLOv5, YOLOX}. Specifically, RTMDet-tiny achieves 41.1\% AP at 1020 FPS with only 4.8M parameters. RTMDet-s yields 44.6\% AP with 819 FPS, surpassing previous state-of-art small models.
When extended to instance segmentation and rotated object detection, RTMDet obtained new state-of-the-art performance on the real-time scenario on both benchmarks, with 44.6\% mask AP at 180 FPS on COCO \texttt{val} set and 81.33\% AP on DOTA v1.0, respectively.

\section{Related Work}
\noindent\textbf{Efficient neural architecture for object detection.} 
Object detection aims to recognize and localize objects in the scene.
For real-time applications, existing works mainly explore anchor-based~\cite{liu2016_ssd, YOLO9000, lin2017_focal} or anchor-free~\cite{tian2019fcos, XingyiZhou2019ObjectsAP} one-stage detectors, instead of two-stage detectors~\cite{girshick2015fast, ren2015faster, cascade_rcnn, pang2019libra}.
To improve the model efficiency, efficient backbone networks and model scaling strategies~\cite{YOLOv4, YoungwanLee2019AnEA, YOLOv7} and enhancement of multi-scale feature~\cite{liu2018_panet, efficientdet, nasfpn, giraffedet, fpg, lin2017_fpn} are explored either by handcrafted design or neural architecture search~\cite{nasfpn, chen2019detnas, wang2019nasfcos, spinenet}.
Recent advances also explore model re-parameterization~\cite{repvgg, YOLOv6, YOLOv7, PPYOLOE} to improve the inference speed after model deployment.
In this paper, we contribute an overall architecture with compatible capacity in the backbone and neck, constructed by a new basic building block with large-kernel depth-wise convolutions toward a more efficient object detector.

\begin{figure*}[t]
    \centering
    \includegraphics[width=0.9\linewidth]{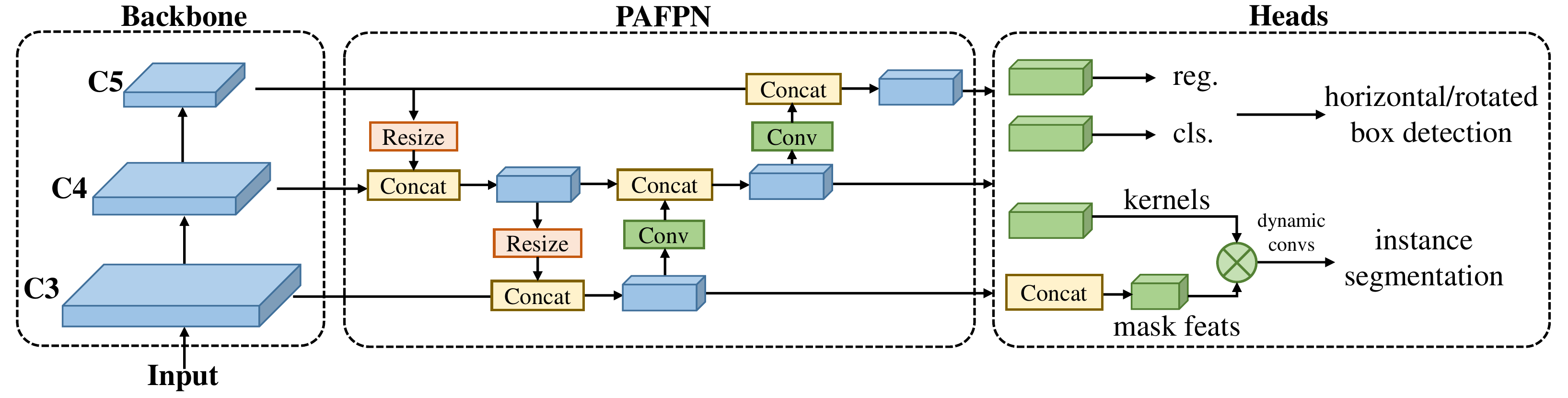}
    \vspace{-10pt}
    \caption{Macro architecture. We use CSP-blocks~\cite{cspnet} with large kernel depth-wise convolution layers to build the backbone. The multi-level features, noted as $C3$, $C4$, and $C5$, are extracted from the backbone and then fused in the CSP-PAFPN, which consists of the same block as the backbone. Then, detection heads with shared convolution weights and separated batch normalization (BN) layers are used to predict the classification and regression results for (rotated) bounding box detection. Extra heads can be added to produce dynamic convolution kernels and mask features for the instance segmentation task.}
    \label{fig:/model_arch}
    \vspace{-12pt}
\end{figure*}

\noindent\textbf{Label assignment for object detection.}
Another dimension to improve the object detector is the design of label assignment and training losses.
Pioneer methods~\cite{ren2015faster, cascade_rcnn, liu2016_ssd, lin2017_focal} use IoU as a matching criterion to compare the ground truth boxes with model predictions or anchors in the label assignment.
Later practices~\cite{ATSS, PAA, tian2019fcos, XingyiZhou2019ObjectsAP} further explore different matching criteria such as object centers~\cite{tian2019fcos, XingyiZhou2019ObjectsAP}.
Auxiliary detection heads are also explored~\cite{NanoDet, YOLOv7} to speed up and stabilize the training.
Inspired by the Hungarian Assignment for end-to-end object detection~\cite{DETR}, dynamic label assignment~\cite{OTA, TOOD, YOLOX} are explored to significantly improve the convergence speed and model accuracy.
Unlike these strategies that use matching cost functions precisely the same as losses, we propose to use soft labels when calculating the matching costs to enlarge the distinction between high and low-quality matches, thereby stabilizing training and accelerating convergence.


\paragraph{Instance segmentation.} 
Instance segmentation aims at predicting the per-pixel mask for each object of interest.
Pioneer methods explore different paradigms to tackle this task, including mask classification~\cite{deepmask, sharpmask}, `Top-Down'~\cite{mask_rcnn, Chen_2019_CVPR}, and `Bottom-Up' approaches~\cite{neven2019instance,kirillov2017instancecut,bai2017deep}.
Recent attempts perform instance segmentation in one stage with~\cite{condinst, bolya2019yolact} or without bounding boxes~\cite{wang2020solo,solov2, knet}.
A representative of these attempts is based on dynamic kernels~\cite{condinst, solov2, knet}, which learn to generate dynamic kernels from either learned parameters~\cite{knet} or dense feature maps~\cite{condinst, solov2} and use them to conduct convolution with mask feature maps.
Inspired by these works, we extend RTMDet by kernel prediction and mask feature heads~\cite{condinst} to conduct instance segmentation.

\paragraph{Rotated object detection.} Rotated object detection aims to predict further the orientation of objects in addition to their locations and categories. 
Based on an existing general object detector (\eg, RetinaNet~\cite{lin2017_focal} or Faster R-CNN~\cite{ren2015faster}), different feature extraction networks are proposed to alleviate the feature misalignment~\cite{han2021redet,yang2021r3det,han2021align} caused by object rotations.
There are also various representations of rotated boxes explored (\eg, Gaussian distribution \cite{yang2021rethinking,yang2021learning} and convex set \cite{guo2021cfa,li2022orep}) to ease the rotated bounding box regression task.
Orthogonal to these methods, this paper only extends a general object detector with minimal modifications (\ie, adding an angle prediction branch and replacing the GIoU~\cite{giou} loss by Rotated IoU Loss~\cite{zhou2019riou}) and reveals that a high-precision general object detector paves the way for high-precision rotated object detection through the model architecture and the knowledge learned on general detection dataset~\cite{lin2014coco}.


\section{Methodology}\label{sec:methods}
In this work, we build a new family of \emph{\textbf{R}eal-\textbf{T}ime \textbf{M}odels for object \textbf{Det}ection}, named \textbf{RTMDet}.
The macro architecture of RTMDet is a typical one-stage object detector (Sec.~\ref{sec:macro}).
We improve the model efficiency by exploring the large-kernel convolutions in the basic building block of backbone and neck, and balance the model depth, width, and resolution accordingly (Sec.~\ref{sec:model}).
We further explore soft labels in dynamic label assignment strategies and a better combination of data augmentations and optimization strategies to improve the model accuracy (Sec.~\ref{sec:training}).
RTMDet is a versatile object recognition framework that can be extended to instance segmentation and rotated object detection tasks with few modifications (Sec.~\ref{sec:generality}).

\subsection{Macro Architecture}\label{sec:macro}

We decompose the macro architecture of a one-stage object detector into the backbone, neck, and head, as shown in Fig.~\ref{fig:/model_arch}.
Recent advances of YOLO series~\cite{YOLOv4, YOLOX} typically adopt CSPDarkNet~\cite{YOLOv4} as the backbone architecture, which contains four stages and each stage is stacked with several basic building blocks (Fig.~\ref{fig:basic_blocks}.a).
The neck takes the multi-scale feature pyramid from the backbone and uses the same basic building blocks as the backbone with bottom-up and top-down feature propogation~\cite{lin2017_fpn, liu2018_panet} to enhance the pyramid feature map.
Finally, the detection head predicts object bounding boxes and their categories based on the feature map of each scale.
Such an architecture generally applies to general and rotated objects, and can be extended for instance segmentation by the kernel and mask feature generation heads~\cite{condinst}.

To fully exploit the potential of the macro architecture, we first study more powerful basic building blocks. Then we investigate the computation bottleneck in the architecture and balance the depth, width, and resolution in the backbone and neck.

\subsection{Model Architecture}\label{sec:model}
\begin{figure}[t]
    \centering
    \includegraphics[width=1.02\linewidth]{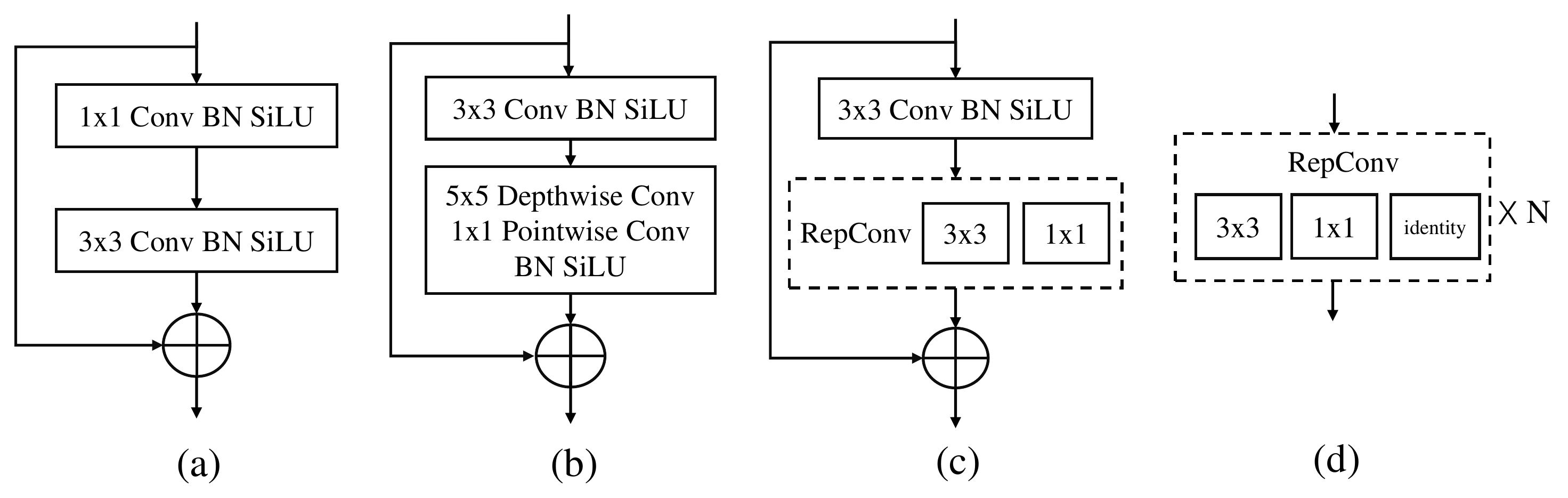}
    \vspace{-12pt}
    \caption{Different basic building blocks. (a) The basic bottleneck block of DarkNet used in~\cite{YOLOv4, YOLOv3, YOLOX, YOLOv5}. (b) The proposed bottleneck block with a large-kernel depth-wise convolution layer. (c) Bottleneck block of PPYOLO-E~\cite{PPYOLOE} that uses re-parameterized convolution. (d) The basic unit of YOLOv6~\cite{YOLOv6}.}
    \label{fig:basic_blocks}
    \vspace{-12pt}
\end{figure}

\paragraph{Basic building block.}
A large effective receptive field in the backbone is beneficial for dense prediction tasks like object detection and segmentation as it helps to capture and model the image context~\cite{receptive_field} more comprehensively.
However, previous attempts (\eg, dilated convolution~\cite{DRN} and non-local blocks~\cite{non_local}) are computationally expensive, limiting their practical use in real-time object detection.
Recent studies~\cite{convnext, replknet} revisit the use of large-kernel convolutions, showing that one can enlarge the receptive field with a reasonable computational cost through depth-wise convolution~\cite{mobilenet}.
Inspired by these findings, we introduce 5$\times$5 depth-wise convolutions in the basic building block of CSPDarkNet~\cite{YOLOv4} to increase the effective receptive fields (Fig.~\ref{fig:basic_blocks}.b). This approach allows for more comprehensive contextual modeling and significantly improves accuracy.

It is noteworthy that some recent real-time object detectors~\cite{YOLOv6, YOLOv7, PPYOLOE} explore re-parameterized 3$\times$3 convolutions~\cite{repvgg} in the basic building block (Fig.~\ref{fig:basic_blocks}.c\&d).
While the re-parameterized 3$\times$3 convolutions is considered a free lunch to improve accuracy during inference, it also brings side effects such as slower training speed and increased training memory.
It also increases the error gap after the model is quantized to lower bits, requiring compensation through re-parameterizing optimizer~\cite{repoptim} and quantization-aware training\cite{YOLOv6}.
Large-kernel depth-wise convolution is a simpler and more effective option for the basic building block compared to re-parameterized 3$\times$3 convolution, as they require less training cost and cause less error gaps after model quantization. 

\paragraph{Balance of model width and depth.}
The number of layers in the basic block also increases due to the additional point-wise convolution following the large-kernel depth-wise convolution (Fig.~\ref{fig:basic_blocks}.b). 
This hinders the parallel computation of each layer and thus decreases inference speed.
To address this issue, we reduce the number of blocks in each backbone stage and moderately enlarging the width of the block to increase the parallelization and maintain the model capacity, which eventually improves inference speed without sacrificing accuracy.

\paragraph{Balance of backbone and neck.}
Multi-scale feature pyramid is essential for object detection to detect objects at various scales. To enhance the multi-scale features, previous approaches either use a larger backbone with more parameters or use a heavier neck~\cite{efficientdet, giraffedet} with more connections and fusions among feature pyramid.
However, these attempts also increase the computation and memory footprints.
Therefore, we adopt another strategy that puts more parameters and computations from backbone to neck by increasing the expansion ratio of basic blocks in the neck to make them have similar capacities, which obtains a better computation-accuracy trade-off.

\paragraph{Shared detection head.}
Real-time object detectors typically utilize separate detection heads~\cite{liu2016_ssd, YOLOv3, YOLOv4, YOLOv5, YOLOX} for different feature scales to enhance the model capacity for higher performance, instead of sharing a detection head across multiple scales~\cite{lin2017_focal, tian2019fcos}.
We compare different design choices in this paper and choose to share parameters of heads across scales but incorporate different Batch Normalization (BN) layers to reduce the parameter amount of the head while maintaining accuracy. BN is also more efficient than other normalization layers such as Group Normalization~\cite{GroupNorm} because in inference it directly uses the statistics calculated in training.

\subsection{Training Strategy}\label{sec:training}
\paragraph{Label assignment and losses.}
To train the one-stage object detector, the dense predictions from each scale will be matched with ground truth bounding boxes through different label assignment strategies~\cite{lin2017_focal, tian2019fcos, TOOD}.
Recent advances typically adopt dynamic label assignment strategies~\cite{DETR, OTA, YOLOX} that use cost functions consistent with the training loss as the matching criterion. However, we find that their cost  calculation have some limitations. Hence, we propose a dynamic soft label assignment strategy based on SimOTA~\cite{YOLOX}, and its cost function is formulated as
\begin{equation}
    C=\lambda_{1} C_{cls} + \lambda_{2} C_{reg} + \lambda_{3} C_{center},
\end{equation}
where $C_{cls}$, $C_{center}$, and $C_{reg}$ correspond to the classification cost, region prior cost, and regression cost, respectively, and $\lambda_{1}=1$, $\lambda_{2}=3$, and $\lambda_{3}=1$ are the weights of these three costs by default. The calculation of the three costs is described below.

Previous methods usually utilize binary labels to compute classification cost $C_{cls}$, which allows a prediction with a high classification score but an incorrect bounding box to achieve a low classification cost and vice versa. To solve this issue, we introduce soft labels in $C_{cls}$ as
\begin{equation}
C_{cls}=CE(P,Y_{soft})\times(Y_{soft}-P)^2.
\end{equation}
The modification is inspired by GFL~\cite{gfl} that uses the IoU between the predictions and ground truth boxes as the soft label $Y_{soft}$ to train the classification branch.
The soft classification cost in assignment not only reweights the matching costs with different regression qualities but also avoids the noisy and unstable matching caused by binary labels.

When using Generalized IoU~\cite{giou} as regression cost, the maximum difference between the best match and the worst match is less than 1. This makes it difficult to distinguish high-quality matches from low-quality matches. To make the match quality of different GT-prediction pairs more discriminative, we use the logarithm of the IoU as the regression cost instead of GIoU used in the loss function, which amplifies the cost for matches with lower IoU values.
The regression cost $C_{reg}$ is calculated by
\begin{equation}
C_{reg}=-log(IoU).
\end{equation} 

For region cost $C_{center}$, we use a soft center region cost instead of a fixed center prior~\cite{ATSS, OTA, YOLOX} to stabilize the matching of the dynamic cost as below
\begin{equation}
C_{center} = \alpha^{\left| x_{pred}-x_{gt} \right| -\beta},
\end{equation}
where $\alpha$ and $\beta$ are hyper-parameters of the soft center region. We set $\alpha=10$, $\beta=3$ by default.
%

\paragraph{Cached Mosaic and MixUp.}
Cross-sample augmentations such as MixUp~\cite{mixup} and CutMix~\cite{CutMix} are widely adopted in recent object detectors~\cite{YOLOv4, YOLOv5,YOLOv6,YOLOv7,YOLOX}.
%
These augmentations are powerful but bring two side effects.
First, at each iteration, they need to load multiple images to generate a training sample, which introduces more data loading costs and slows the training.
Second, the generated training sample is `noisy' and may not belong to the real distribution of the dataset, which affects the model learning~\cite{YOLOX}.

We improve MixUp and Mosaic with the caching mechanism that reduces the demand for data loading.
By utilizing cache, the time cost of mixing images in the training pipeline can be significantly reduced to the level of processing a single image.
The cache operation is controlled by the cache length and popping method. 
A large cache length and random popping method can be regarded as equivalent to the original non-cached MixUp and Mosaic operations.
Meanwhile, a small cache length and First-In-First-Out (FIFO) popping method can be seen as similar to the repeated augmentation~\cite{berman2019multigrain}, allowing for the mixing of the same image with different data augmentation operations in the same or contiguous batches.
%

\paragraph{Two-stage training.}
To reduce the side effects of `noisy' samples by strong data augmentations, 
YOLOX~\cite{YOLOX} explored a two-stage training strategy, where the first stage uses strong data augmentations, including Mosaic, MixUp, and random rotation and shear, and the second stage use weak data augmentations, such as random resizing and flipping.
As the strong augmentation in the initial training stage includes random rotation and shearing that cause misalignment between inputs and the transformed box annotations, YOLOX adds the L1 loss to fine-tune the regression branch in the second stage.
To decouple the usage of data augmentation and loss functions, we exclude these data augmentations and increase the number of mixed images to 8 in each training sample in the first training stage of 280 epochs to compensate for the strength of data augmentation. 
In the last 20 epochs, we switch to Large Scale Jittering (LSJ)~\cite{scp}, allowing for fine-tuning of the model in a domain that is more closely aligned with the real data distributions.
To further stabilize the training, we adopt AdamW~\cite{ADAMW} as the optimizer, which is rarely used in convolutional object detectors but is a default for vision transformers~\cite{vit}.
%


\subsection{Extending to other tasks}\label{sec:generality}

\begin{figure}[t]
    \centering
    \includegraphics[width=0.9\linewidth]{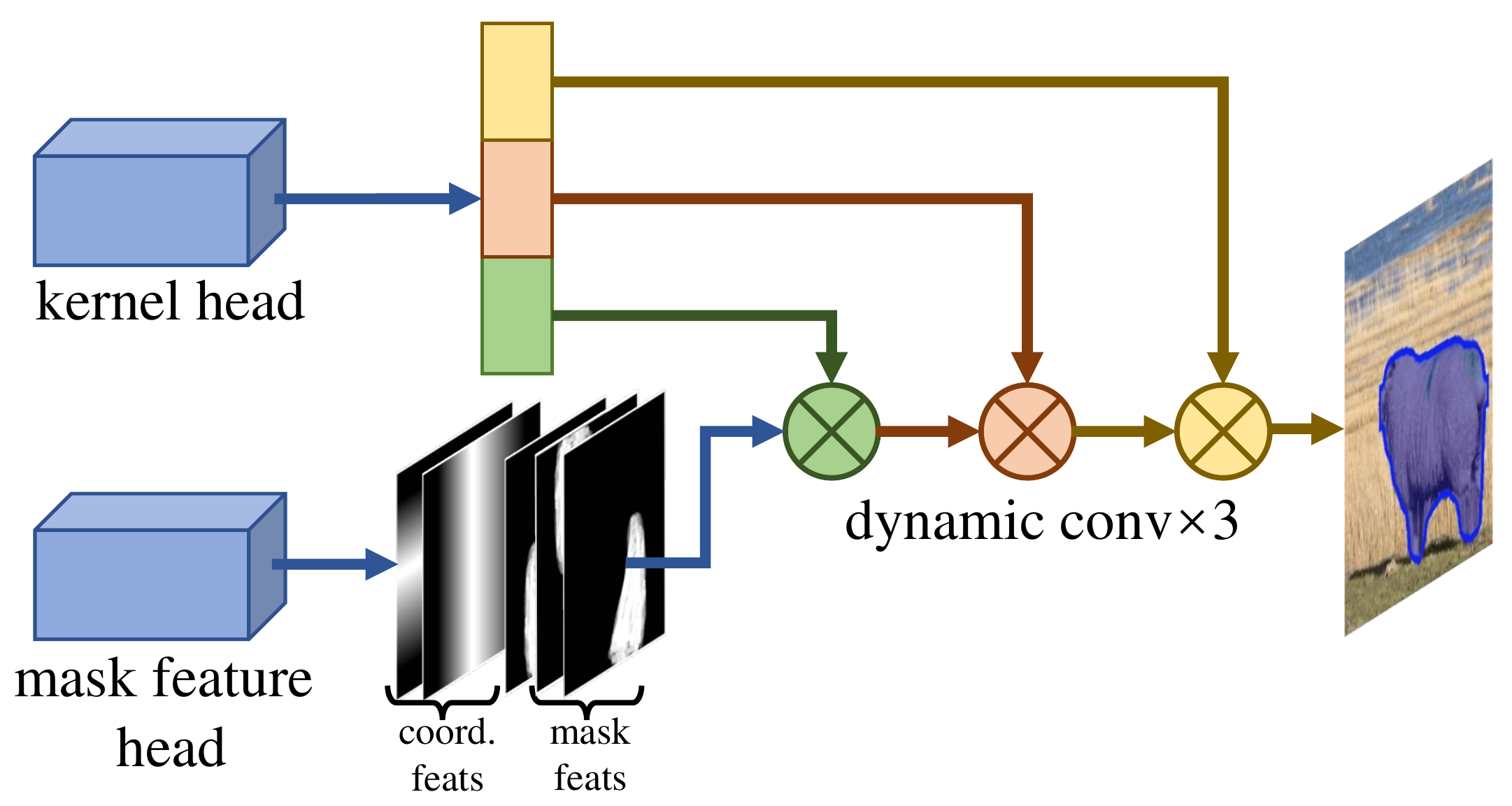}
    \vspace{-7pt}
    \caption{Instance segmentation branch in RTMDet-Ins. The mask feature head has 4 convolution layers and predicts mask features of 8 channels~\cite{condinst} from the multi-level features extracted from neck. Two relative coordinate features are concatenated with the mask features to generate instance masks. The kernel head predicts a 169-dimensional vector for each instance. The vector is divided into three parts (lengths are 88, 72, and 9 respectively), which are used to form the kernels of three dynamic convolution layers.}
    \label{fig:dynamic_conv}
    \vspace{-12pt}
\end{figure}

\paragraph{Instance segmentation.} We enable RTMDet for instance segmentation with a simple modification, denoted as RTMDet-Ins.
As illustrated in Figure~\ref{fig:dynamic_conv}, based on RTMDet, an additional branch is added, consisting of a kernel prediction head and a mask feature head, similar to CondInst~\cite{condinst}.
The mask feature head comprises 4 convolution layers that extract mask features with 8 channels from multi-level features. The kernel prediction head predicts a 169-dimensional vector for each instance, which is decomposed into three dynamic convolution kernels to generate instance segmentation masks through interaction with the mask features and coordinate features.
To further exploit the prior information inherent in the mask annotations, we use the mass center of the masks when calculating the soft region prior in the dynamic label assignment instead of the box center. 
We use dice loss~\cite{vnet} as the supervision for the instance masks following typical conventions.

\paragraph{Rotated object detection.} Due to the inherent similarity between rotated object detection and general (horizontal) object detection, it only takes 3 steps to adapt RTMDet to a rotated object detector, noted as RTMDet-R: (1) add a 1$\times$1 convolution layer at the regression branch to predict the rotation angle; (2) modify the bounding box coder to support rotated boxes; (3) replace the GIoU loss with Rotated IoU loss. The highly optimized model architecture of RTMDet guarantees high performance of RTMDet-R on the rotated object detection tasks.
Moreover, as RTMDet-R shares most parameters of RTMDet, the model weights of RTMDet pre-trained on general detection dataset (\eg, COCO dataset) can serve as a good initialization for rotated object detection.

\section{Experiments}\label{sec:Experiments}
\subsection{Implementation Details}
\paragraph{Object detection and instance segmentation.}
We conduct experiments on COCO dataset~\cite{lin2014coco}, which contains about 118K images in the \texttt{train2017} set and 5K images in the \texttt{val2017} set for training and validation, respectively.
For ablation studies, we trained our models on the \texttt{train2017} set for 300 epochs and validated them on the \texttt{val2017} set. The hyper-parameters are in Table~\ref{tab:training_setting}. All our object detection and instance segmentation models are trained on 8 NVIDIA A100 GPUs. We evaluate the model performance on object detection and instance segmentation by bbox AP and mask AP~\cite{lin2014coco}, respectively. 

During the testing of object detection, the score threshold for filtering bounding boxes is set to 0.001 before non-maximum suppression (NMS), and the top 300 boxes are kept for validation. This setting is consistent with previous studies~\cite{YOLOv5, YOLOv6, YOLOv7} for a fair comparison. However, to accelerate the metric computation during ablation studies, we set the score threshold to 0.05 and the number of kept results to 100, which may degrade the accuracy by about 0.3\% AP.

\paragraph{Rotated object detection.} 
We conduct experiments on DOTA dataset \cite{xia2018dota} which contains 2.8K aerial images and 188K instances obtained from different sensors with multiple resolutions.
The hyper-parameters are in Table~\ref{tab:training_setting}. For single-scale training and testing, we crop the original images into 1024$\times$1024 patches with an overlap of 256 pixels. For multi-scale training and testing, the original images are resized with the scale of 0.5, 1.0 and 1.5 and then cropped into 1024$\times$1024 patches with an overlap of 500 pixels. 
Most of the rotated object detectors are trained by 1 NVIDIA V100 GPU except that the large model uses 2 NVIDIA V100 GPUs.
For the evaluation metric, we adopt the same mAP calculation as that in PASCAL VOC2007 \cite{mark2010voc} but use rotated IoU to calculate the matched objects.


\begin{table}[h]
  \centering
    \caption{\small{\textbf{Training settings for object detection, instance segmentation and rotate object detection.}}}\label{tab:training_setting}
\vspace{-6pt}
 \tablestyle{1pt}{1.2}
  \scalebox{0.9}{
  \begin{tabular}{x{86} | x{82} | x{80}}
    config & Object detection and instance segmentation & Rotate object detection\\
    \shline
    optimizer                       & AdamW \cite{ADAM} & AdamW\\
    base learning rate              & 0.004 & 0.00025\\
    weight decay                    & 0.05 (0 for bias and norm \cite{bag_of_tricks}) & 0.05 (0 for bias and norm)\\
    optimizer momentum              & 0.9 & 0.9\\
    batch size                      & 256 & 8\\
    learning rate schedule          & Flat-Cosine & Flat-Cosine\\
    training epochs                 & 300 & 36\\
    warmup iterations               & 1000 & 1000\\
    input size                      & 640 \x 640 & 1024 \x 1024\\
    augmentation                    & cached Mosaic and MixUp (first 280 epochs); LSJ \cite{scp, detectron2} (last 20 epochs) & random Flip and Rotate \\
    EMA decay                       & 0.9998 & 0.9998
    \end{tabular}
    }
      \vspace{-3pt}
\end{table}

\begin{table*}[t]
  \centering
  \vspace{-5pt}
    \caption{\small{\textbf{Comparison of RTMDet with previous practices} on the number of parameters, FLOPS, latency, and accuracy on COCO \texttt{val2017} set. For a fair comparison, all models are trained for 300 epochs without using extra detection data or knowledge distillation. The inference speeds of all models are measured in the same environment. (LB) means LetterBox resize proposed in \cite{YOLOv5}. The results of the proposed RTMDet are marked in gray. The best results are in bold}}\label{tab:overall}
 \tablestyle{1pt}{1.2}
  \scalebox{0.95}{
  \begin{tabular}{ c | x{60}x{50}x{50}x{55}|x{35}x{43}}
    Model       & Input shape & Params(M)~$\downarrow$ & FLOPs(G)~$\downarrow$  & Latency(ms)~$\downarrow$ & AP(\%) ~$\uparrow$           & AP50(\%) ~$\uparrow$ \\
    \shline
    YOLOv5-n~\cite{YOLOv5}    & 640(LB)     & 1.9    & 2.3   & 1.51    & 28.0            & 45.7 \\
    YOLOX-tiny~\cite{YOLOX}   & 416$\times$416     & 5.1   & 3.3    & 0.82    & 32.8          & 50.3 \\
    YOLOv6-n~\cite{YOLOv6}     & 640(LB)     & 4.3    & 5.6    & 0.79    & 35.9          & 51.2 \\
    YOLOv6-tiny & 640(LB)     & 9.7   & 12.5   & 0.86    & 40.3          & 56.6 \\
    \ours{RTMDet-tiny} & \ours{640$\times$640}     & \ours{4.8}    & \ours{8.1}    & \ours{0.98}    & \ours{\textbf{41.1}} & \ours{\bd{57.9}} \\
    \shline
    YOLOv5-s     & 640(LB)     & 7.2    & 8.3   & 1.63    & 37.4          & 56.8 \\
    YOLOX-s     & 640$\times$640     & 9.0    & 13.4   & 1.20    & 40.5          & 59.3 \\
    YOLOv6-s    & 640(LB)     & 17.2   & 22.1   & 0.92    & 43.5          & 60.4 \\
    PPYOLOE-s~\cite{PPYOLOE}   & 640$\times$640     & 7.9   & 8.7   & 1.34       & 43.0          & 59.6    \\
    \ours{RTMDet-s}    & \ours{640$\times$640}     & \ours{8.99}   & \ours{14.8}   & \ours{1.22}    & \ours{\bd{44.6}} & \ours{\bd{61.9}} \\
    \shline
    YOLOv5-m     & 640(LB)     & 21.2   & 24.5   & 1.89    & 45.4          & 64.1 \\
    YOLOX-m     & 640$\times$640     & 25.3   & 36.9   & 1.68    & 46.9          & 65.6 \\
    YOLOv6-m     & 640(LB)    & 34.3   & 41.1   & 1.21    & 48.5          & -    \\
    PPYOLOE-m   & 640$\times$640     & 23.4   & 25.0   & 1.75   & 49.0           & 65.9    \\
    \ours{RTMDet-m}    & \ours{640$\times$640}     & \ours{24.7}  & \ours{39.3}  & \ours{1.62}    & \ours{\textbf{49.4}} & \ours{\bd{66.8}}\\
    \shline
    YOLOv5-l     & 640(LB)     & 46.5   & 54.6   & 2.46    & 49.0            & 67.3 \\
    YOLOX-l     & 640$\times$640     & 54.2   & 77.8   & 2.19    & 49.7          & 68.0 \\
    YOLOv6-l     & 640(LB)     & 58.5   & 72.0     & 1.91    & 51.0            & -    \\
    YOLOv7~\cite{YOLOv7}     & 640(LB)     & 36.9   & 52.4     & 2.63    & 51.2            & -    \\
    PPYOLOE-l   & 640$\times$640     & 52.2   & 55.0     & 2.57    & 51.4          & 68.6    \\
    \ours{RTMDet-l}    & \ours{640$\times$640}     & \ours{52.3}   & \ours{80.2}  & \ours{2.40}    & \ours{\bd{51.5}}          & \ours{\bd{68.8}} \\
    \shline
    YOLOv5-x     & 640(LB)     & 86.7   & 102.9  & 2.92    & 50.7          & 68.9 \\
    YOLOX-x     & 640$\times$640     & 99.1   & 141.0 & 2.98    & 51.1          & 69.4 \\
    PPYOLOE-x   & 640$\times$640     & 98.4   & 103.3  & 3.07       & 52.3          & 69.5    \\
    \ours{RTMDet-x}    & \ours{640$\times$640}     & \ours{94.9}  & \ours{141.7} & \ours{3.10}    & \ours{\textbf{52.8}} & \ours{\bd{70.4}}
    \end{tabular}
    }
\end{table*}

\paragraph{Benchmark settings.} 
The latency of all models is tested in the half-precision floating-point format (FP16) on an NVIDIA 3090 GPU with TensorRT 8.4.3 and cuDNN 8.2.0. The inference batch size is 1. 

\subsection{Benchmark Results}
\paragraph{Object detection.}
We compare RTMDet with previous real-time object detectors including YOLOv5~\cite{YOLOv5}, YOLOX~\cite{YOLOX}, YOLOv6~\cite{YOLOv6}, YOLOv7~\cite{YOLOv7}, and PPYOLOE~\cite{PPYOLOE}.
For a fair comparison, all models are trained on 300 epochs without distillation nor pruning and the time of Non-Maximum Suppression (NMS) is not included in the latency calculation.

As shown in Table~\ref{tab:overall} and Fig.~\ref{fig:compare_params} (a), RTMDet achieves
a better parameter-accuracy trade-off than previous methods. RTMDet-tiny achieves 41.1\% AP with only 4.8M parameters, surpassing other models with a similar size by more than 5\% AP. RTMDet-s has a higher accuracy with only half of the parameters and computation costs of YOLOv6-s. RTMDet-m and RTMDet-l also achieve excellent results in similar class models, with 44.6\% and 49.4\% AP respectively. RTMDet-x yields 52.8\% AP with 300+FPS, outperforming the current mainstream detectors.
It is worth noting that both ~\cite{YOLOv5} and ~\cite{YOLOv7} use mask annotation to refine the bounding boxes after data augmentation, resulting in a gain of about 0.3\% AP. We achieved superior results without relying on additional information beyond box annotation.

\paragraph{Instance segmentation.}
To evaluate the superiority of our label assignment strategy and loss, we first compare RTMDet-Ins with conventional methods using the standard ResNet50-FPN~\cite{lin2017_fpn} backbone and the classic multi-scale 3x schedule~\cite{mmdetection, detectron2}.
We adopt an auxiliary semantic segmentation head for faster convergence speed and a fair comparison with CondInst~\cite{condinst}.
RTMDet outperforms CondInst by 1.5\% mask AP (the first row in Table~\ref{tab:rtmdet_ins}).
However, we do not use the semantic segmentation branch when training RTMDet from scratch with heavy data augmentation because the auxiliary branch brings marginal improvements.

Finally, we trained RTMDet-Ins tiny/s/m/l/x on the COCO dataset using the same data augmentation and optimization hyper-parameters as RTMDet for 300 epochs. 
RTMDet-Ins-x achieves 44.6\% mask AP, surpasses the previous best practice YOLOv5-seg-x~\cite{YOLOv5} by 3.2\% AP, and still runs in real-time (second row in Table~\ref{tab:rtmdet_ins}).

\begin{table*}[th]
  \centering
  \vspace{-5pt}
    \caption{\small{\textbf{Comparison of RTMDet-Ins with previous instance segmentation methods} on the number of parameters, FLOPS, latency, and accuracy on COCO \texttt{val2017} set. (LB) means LetterBox resize proposed in \cite{YOLOv5}. The results of the proposed RTMDet-Ins are marked in gray. The best results are in bold. Different from the object detection model, box NMS and post-processing of top-100 masks are included in the speed measurement}}\label{tab:rtmdet_ins}
 \tablestyle{1pt}{1.2}
  \scalebox{0.95}{\begin{tabular}{ c | x{60}x{50}x{50}x{50}x{55}|x{50}x{55}}
        Model            & Input shape & Epochs & Params(M)~$\downarrow$ & FLOPs(G)~$\downarrow$ & Latency(ms)~$\downarrow$ & Box AP(\%)~$\uparrow$        & Mask AP(\%)~$\uparrow$       \\ \shline
        SparseInst-R50~\cite{Cheng2022SparseInst}      & 640-853   & 147      & 31.6   & 99.1  &  -          & -          & 34.2          \\
        SOLOv2-R50-FPN~\cite{solov2}      & 800-1333   & 36      & 46.4   & 253.5  &  -          & -       & 37.5          \\
        CondInst-R50-FPN~\cite{condinst}      & 800-1333   & 36      & 33.9   & 240.8  &  -          & 42.6       & 38.2          \\
        Cascade-R50-FPN~\cite{cascade_rcnn}   & 800-1333   & 36      & 77.1   & 403.6  &  -          & 44.3       & 38.5          \\
        \ours{RTMDet-Ins-R50-FPN}& \ours{800-1333} & \ours{36} & \ours{35.9}  & \ours{295.2} & \ours{-} & \ours{\bd{45.3}} & \ours{\bd{39.7}}          \\  \shline
        YOLOv5n-seg~\cite{YOLOv5}      & 640(LB)   & 300      & 2.0         & 3.6      &    1.65        & 27.6          & 23.4          \\
        YOLOv5s-seg      & 640(LB)   & 300      & 7.6       & 13.2     &    1.90    & 37.6          & 31.7          \\
        YOLOv5m-seg      & 640(LB)   & 300      & 22        & 35.4     &    2.71      & 45.0          & 37.1          \\
        YOLOv5l-seg      & 640(LB)   & 300      & 47.9      & 73.9     &    3.44     & 49.0          & 39.9          \\
        YOLOv5x-seg      & 640(LB)   & 300      & 88.8      & 132.9    &    5.10    & 50.7          & 41.4          \\
        \ours{RTMDet-Ins-tiny}& \ours{640$\times$640} & \ours{300} & \ours{5.6} & \ours{11.8} & \ours{1.70} & \ours{40.5}  & \ours{35.4} \\
        \ours{RTMDet-Ins-s}& \ours{640$\times$640} & \ours{300} & \ours{10.2} & \ours{21.5} & \ours{1.93} & \ours{44.0}  & \ours{38.7} \\
        \ours{RTMDet-Ins-m}& \ours{640$\times$640} & \ours{300} & \ours{27.6} & \ours{54.1} & \ours{2.69} & \ours{48.8}  & \ours{42.1} \\
        \ours{RTMDet-Ins-l}& \ours{640$\times$640} & \ours{300} & \ours{57.4} & \ours{106.6} & \ours{3.68} & \ours{51.2}  & \ours{43.7} \\
        \ours{RTMDet-Ins-x}& \ours{640$\times$640} & \ours{300} & \ours{102.7} & \ours{182.7} & \ours{5.31} & \ours{\bd{52.4}}  & \ours{\bd{44.6}} \\
    \end{tabular}}
\end{table*}

\begin{table*}[ht]
\centering
  \vspace{-3pt}
    \caption{\small{\textbf{Comparison of RTMDet-R with previous rotated object detection methods} on the number of parameters, FLOPs, latency, and accuracy on DOTA-v1.0 test set. IN and COCO denote ImageNet pretraining and COCO pretraining. MAE means MAE unsupervised pretraining  \cite{he2021mae} on the MillionAID \cite{long2021creating}. R50 and X50 denote ResNet-50 and ResNeXt-50 (likewise for R101, R152 and X101). Re50 denotes ReResNet-50, RVSA denotes RVSA-ViTAE-B and CRN denotes CSPRepResNet. MS means multi-scale training and testing. DOTA-v1.0 has 15 different object categories: plane (PL), baseball diamond (BD), bridge (BR), ground track field (GTF), small vehicle (SV), large vehicle (LV), ship (SH), tennis court (TC), basketball court (BC), storage tank (ST), soccer ball field (SBF), roundabout (RA), harbor (HA), swimming pool (SP), and helicopter (HC). The AP of each category is listed. The bold fonts indicate the best performance. The results of the proposed RTMDet-R are marked in gray}}\label{tab:dota}
\resizebox{\textwidth}{!}{
\begin{tabular}{c|c|c|c|c|ccccccccccccccc}
Method & Pretrain & Backbone & MS &  mAP(\%) &  PL &  BD &  BR &  GTF &  SV &  LV &  SH &  TC &  BC &  ST &  SBF &  RA &  HA &  SP &  HC \\ \hline
\textbf{\emph{Anchor-based Methods}}&  &  &  &  &  &  &  &  &  &  &  &  &  &  &  &  &  &  &  \\

RoI Trans. \cite{ding2018learning} & IN & R101 \cite{He_2016} & $\checkmark$ & 69.56 & 88.64 & 78.52 & 43.44 & 75.92 & 68.81 & 73.68 & 83.59 & 90.74 & 77.27 & 81.46 & 58.39 & 53.54 & 62.83 & 58.93 &  47.67   \\

Gliding Vertex \cite{xu2020gliding} & IN & R101 & $\checkmark$ & 75.02 & 89.64 & 85.00 & 52.26 & 77.34 & 73.01 & 73.14 & 86.82 & 90.74 & 79.02 & 86.81 & 59.55 & 70.91 & 72.94 & 70.86 & 57.32  \\

CSL \cite{yang2020arbitrary} & IN & R152 & $\checkmark$ & 76.17  & \bd{90.25} & \bd{85.53} & 54.64 & 75.31 & 70.44 & 73.51 & 77.62 & 90.84 & 86.15 & 86.69 & 69.60 &  68.04 & 73.83 & 71.10 & 68.93\\

R$^{3}$Det \cite{yang2021r3det} & IN & R152 & $\checkmark$ & 76.47 & 89.80 & 83.77 & 48.11 &  66.77 & 78.76 & 83.27 & 87.84 & 90.82 &  85.38 & 85.51 & 65.57 & 62.68 &  67.53 & 78.56 & 72.62  \\

DCL  \cite{yang2021dense} & IN & R152 & $\checkmark$ & 77.37 & 89.26 & 83.60 & 53.54 & 72.76 & 79.04 &  82.56 & 87.31 & 90.67 &  86.59 & 86.98 & 67.49 &  66.88 & 73.29 & 70.56 & 69.99  \\

S$^{2}$ANet \cite{han2021align} & IN & R50 & $\checkmark$ & 79.42 & 88.89 & 83.60 & 57.74 & 81.95 & 79.94 & 83.19 & \bd{89.11} & 90.78 & 84.87 & 87.81 & 70.30 &  68.25 & 78.30 & 77.01 &  69.58  \\

ReDet \cite{han2021redet} & IN & Re50 \cite{han2021redet} & $\checkmark$ & 80.10 & 88.81 & 82.48 & 60.83 & 80.82 & 78.34 & 86.06 & 88.31 & 90.87 & \bd{88.77} & 87.03 & 68.65 & 66.90 & 79.26 & 79.71 & 74.67   \\

GWD \cite{yang2021rethinking} & IN & R152 & $\checkmark$ & 80.23 & 89.66 & 84.99 & 59.26 & 82.19 & 78.97 & 84.83 & 87.70 & 90.21 & 86.54 & 86.85 & \bd{73.47} & 67.77 & 76.92 & 79.22 & 74.92   \\

KLD \cite{yang2021learning} & IN & R152 & $\checkmark$ & 80.63 & 89.92 & 85.13 & 59.19 & 81.33 & 78.82 & 84.38 & 87.50 & 89.80 & 87.33 & 87.00 & 72.57 & 71.35 & 77.12 & 79.34 & 78.68   \\

Oriented RCNN \cite{xie2021oriented} & IN & R50 & $\checkmark$ & 80.87 & 89.84 & 85.43 & 61.09 & 79.82 & 79.71 & 85.35 & 88.82 & 90.88 & 86.68 & 87.73 & 72.21 & 70.80 & 82.42 & 78.18 & 74.11  \\

RoI Trans. + KFIoU \cite{yang2022kfiou} & IN & Swin-tiny 
 \cite{swin} & $\checkmark$ & 80.93 & 89.44 & 84.41 & \bd{62.22} & \bd{82.51} & 80.10 & \bd{86.07} & 88.68 & \bd{90.90} & 87.32 & \bd{88.38} & 72.80 & \bd{71.95} & 78.96 & 74.95 & 75.27  \\

Oriented RCNN  & MAE & RVSA \cite{wang2022vitrvsa} & $\checkmark$ & \bd{81.18} & 89.40 & 83.94 & 59.76 & 82.10 & \bd{81.73} & 85.32 & 88.88 & 90.86 & 85.69 & 87.65 & 63.70 & 69.94 & \bd{84.72} & \bd{84.16} & \bd{79.90}  \\

\hline
\textbf{\emph{Anchor-free Methods}}&  &  &  &  &  &  &  &  &  &  &  &  &  &  &  &  &  &  &  \\

CFA \cite{guo2021beyond} & IN & R152 & $\checkmark$ & 76.67 & 89.08 &  83.20 & 54.37 & 66.87 & 81.23 & 80.96 & 87.17 & 90.21 & 84.32 & 86.09 &  52.34 & 69.94 & 75.52 & 80.76 & 67.96  \\

DAFNe \cite{lang2021dafne} & IN & R101 & $\checkmark$ & 76.95 & 89.40 & 86.27 & 53.70 & 60.51 & 82.04 & 81.17 & 88.66 & 90.37 & 83.81 & 87.27 & 53.93 & 69.38 & 75.61 & 81.26 & 70.86  \\

SASM \cite{hou2022shape} & IN & RX101 \cite{Xie_2017} & $\checkmark$ & 77.19 & 88.41 & 83.32 & 54.00 & 74.34 & 80.87 & 84.10 & 88.04 & 90.74 & 82.85  & 86.26 & 63.96 & 66.78 & 78.40 & 73.84 & 61.97 \\

Oriented RepPoints \cite{li2022orep} & IN & Swin-tiny &  & 77.63 & 89.11 & 82.32 & 56.71 & 74.95 & 80.70 & 83.73 & 87.67 & 90.81 & 87.11 & 85.85 & 63.60 & 68.60 & 75.95 & 73.54 & 63.76  \\

PPYOLOE-R-s & IN & CRN-s \cite{PPYOLOE} &  & 73.82 & 88.80 & 79.24 & 45.92 & 66.88 & 80.41 & 82.95 & 88.20 & 90.61 & 82.91 & 86.37 & 55.80 & 64.11 & 65.09 & 79.50 & 50.43 \\
PPYOLOE-R-s & IN & CRN-s & $\checkmark$ & 79.42 & 88.93 & 83.95 & 56.60 & 79.40 & 82.57 & 85.89 & 88.64 & 90.87 & 87.82 & 87.54 & 68.94 & 63.46 & 76.66 & 79.19 & 70.87 \\

PPYOLOE-R-m & IN & CRN-m &  & 77.64 & 89.23 & 79.92 &  51.14 & 72.94 &  81.86 &  84.56 &  88.68 &  90.85 & 86.85 & 87.48 & 59.16 & 68.34 &  73.78 & 81.72 & 68.10 \\
PPYOLOE-R-m & IN & CRN-m & $\checkmark$ & 79.71 & 88.63 & 84.45 & 56.27 & 79.12 & \bd{83.52} & \bd{86.16} & 88.77 & 90.81 & 88.01 & \bd{88.39} & 70.41 & 61.44 & 77.65 & 77.70 & 74.30 \\

PPYOLOE-R-l & IN & CRN-l &  & 78.14 & 89.18 & 81.00 & 54.01 & 70.22 & 81.85 & 85.16 & 88.81 & 90.81 &  86.99 & 88.01 & 62.87 & 67.87 & 76.56 & 79.13 & 69.65 \\
PPYOLOE-R-l & IN & CRN-l & $\checkmark$ & 80.02 & 88.40 & 84.75 & 58.91 & 76.35 & 83.13 & 86.10 &  88.79 &  90.87 & 88.74 & 87.71 & 67.71 & 68.44 & 77.92 & 76.17 & 76.35 \\

PPYOLOE-R-x & IN & CRN-x &  & 78.28 & \bd{89.49} & 79.70 & 55.04 & 75.59 & 82.40 & 85.20 & 88.35 & 90.76 & 85.69 & 87.70 & 63.17 & 69.52 & 77.09 & 75.08 &  69.38 \\
PPYOLOE-R-x & IN & CRN-x & $\checkmark$ & 80.73 & 88.45 & 84.46 & \bd{60.57} & 77.70 & 83.34 & 85.36 & \bd{88.97} & 90.78 & 88.53 & 87.47 & 69.26 & 65.96 & 77.86 & 81.36 & \bd{80.93} \\

\ours{RTMDet-R-tiny} & \ours{IN} & \ours{Ours-tiny} & \ours{ } & \ours{75.36} & \ours{89.21} & \ours{80.03} & \ours{47.88} & \ours{69.73} & \ours{82.05} & \ours{83.33} & \ours{88.63} & \ours{\bd{90.91}} & \ours{86.31} & \ours{86.85} & \ours{59.94} & \ours{62.30} & \ours{74.23} & \ours{71.97} & \ours{57.03} \\
\ours{RTMDet-R-tiny} & \ours{IN} & \ours{Ours-tiny} & \ours{$\checkmark$} & \ours{79.82} & \ours{87.89} & \ours{85.70} & \ours{55.83} & \ours{81.28} & \ours{81.47} & \ours{85.12} & \ours{88.91} & \ours{90.88} & \ours{88.15} & \ours{87.96} & \ours{67.29} & \ours{68.59} & \ours{77.71} & \ours{80.50} & \ours{69.96} \\

\ours{RTMDet-R-s} & \ours{IN} & \ours{Ours-s} & \ours{ } & \ours{76.93} & \ours{89.18} & \ours{80.45} & \ours{52.09} & \ours{71.35} & \ours{81.55} & \ours{84.05} & \ours{88.79} & \ours{90.89} & \ours{87.83} & \ours{86.98} & \ours{59.58} & \ours{62.28} & \ours{75.90} & \ours{81.96} & \ours{61.04} \\
\ours{RTMDet-R-s} & \ours{IN} & \ours{Ours-s} & \ours{$\checkmark$}& \ours{79.98}  & \ours{88.16} & \ours{86.09} & \ours{56.80} & \ours{78.79} & \ours{80.62} & \ours{85.06} & \ours{88.64} & \ours{90.82} & \ours{86.90} & \ours{86.70} & \ours{66.23} & \ours{70.22} & \ours{78.17} & \ours{81.71} & \ours{74.58} \\

\ours{RTMDet-R-m} & \ours{IN} & \ours{Ours-m} & \ours{ }  & \ours{78.24} & \ours{89.17} & \ours{84.65} & \ours{53.92} & \ours{74.67} & \ours{81.48} & \ours{83.99} & \ours{88.71} & \ours{90.85} & \ours{87.43} & \ours{87.20} & \ours{59.39} & \ours{66.68} & \ours{77.71} & \ours{\bd{82.40}} & \ours{65.28}\\
\ours{RTMDet-R-m} & \ours{IN} & \ours{Ours-m} & \ours{$\checkmark$} & \ours{80.26} & \ours{87.10} & \ours{85.83} & \ours{56.30} & \ours{80.28} & \ours{80.04} & \ours{84.67} & \ours{88.22} & \ours{90.88} & \ours{88.49} & \ours{87.57} & \ours{70.74} & \ours{69.99} & \ours{78.35} & \ours{80.88} & \ours{74.53} \\

\ours{RTMDet-R-l} & \ours{IN} & \ours{Ours-l} & \ours{ } & \ours{78.85} & \ours{89.43} & \ours{84.21} & \ours{55.20} & \ours{75.06} & \ours{80.81} & \ours{84.53} & \ours{\bd{88.97}} & \ours{90.90} & \ours{87.38} & \ours{87.25} & \ours{63.09} & \ours{67.87} & \ours{78.09} & \ours{80.78} & \ours{69.13} \\
\ours{RTMDet-R-l} & \ours{IN} & \ours{Ours-l} & \ours{$\checkmark$} & \ours{80.54} & \ours{88.36} & \ours{84.96} & \ours{57.33} & \ours{80.46} & \ours{80.58} & \ours{84.88} & \ours{88.08} & \ours{90.90} & \ours{86.32} & \ours{87.57} & \ours{69.29} & \ours{70.61} & \ours{78.63} & \ours{80.97} & \ours{79.24} \\
\ours{RTMDet-R-l} & \ours{COCO} & \ours{Ours-l} & \ours{$\checkmark$} & \ours{\bd{81.33}} & \ours{88.01} & \ours{\bd{86.17}} & \ours{58.54} & \ours{\bd{82.44}} & \ours{81.30} & \ours{84.82} & \ours{88.71} & \ours{90.89} & \ours{\bd{88.77}} & \ours{87.37} & \ours{\bd{71.96}} & \ours{\bd{71.18}} & \ours{\bd{81.23}} & \ours{81.40} & \ours{77.13} \\

\end{tabular}}
\end{table*}

\paragraph{Rotated object detection.} We compare RTMDet-R with previous state of the arts on the DOTA v1.0 dataset as shown in Table~\ref{tab:dota}. With single-scale training and testing, RTMDet-R-m and RTMDet-R-l achieve 78.24\% and 78.85\% mAP, respectively, which outperforms almost all previous methods. With multi-scale training and testing, RTMDet-R-m and RTMDet-R-l further acheives 80.26\% and 80.54\% mAP, respectively. Moreover, RTMDet-R-l (COCO pretraining) sets a new record (81.33\% mAP) on the DOTA-v1.0 dataset.
RTMDet-R also consistently outperforms PPYOLOE-R in all regimes of model sizes with much simpler modifications.
Note that RTMDet-R avoids special operators in the architecture to achieve high precision, which makes it can be easily deployed on various hardware.
We also compare RTMDet-R with other methods on HRSC2016 \cite{liu2016hrsc} and DOTA-v1.5 datasets in the appendix, and RTMDet-R also achieves superior performance.

\begin{table*}[t]
    \centering
    \caption{\textbf{Ablation studies of model architecture} on COCO \texttt{val2017} set. The proposed setting is marked in gray
    }\label{table:ablation_repr}
    \centering
    \begin{minipage}[t]{.48\linewidth}
      \centering
      \subcaption{\small{Speed-accuracy trade-off of kernel size}}\label{tab:ablation:kernel_size}
      \vspace{-3pt}
     \tablestyle{1pt}{1.2}
      \scalebox{0.95}{\begin{tabular}{ c | x{40}x{40}x{40}|x{40}}
        Kernel Size & Params.~$\downarrow$ & GFLOPs~$\downarrow$ & Latency~$\downarrow$ & AP(\%)~$\uparrow$ \\
        \shline
        3$\times$3        & 50.80M  &  79.61G  & 2.10ms & 50.0 \\
        \ours{5$\times$5} & \ours{50.92M}  &  \ours{79.70G}  & \ours{2.11ms} & \ours{50.9} \\
        7$\times$7        & 51.10M  &  80.34G  & 2.73ms & 51.1 \\
        \end{tabular}}

    \end{minipage}
    \begin{minipage}[t]{.48\linewidth}
      \centering
      \subcaption{\small{Speed-accuracy trade-off of the number of blocks}}\label{tab:ablation:num_blocks}
      \vspace{-3pt}
     \tablestyle{1pt}{1.2}
      \scalebox{0.95}{\begin{tabular}{ c | x{40}x{40}x{40}|x{40}}
        Num. Blocks & Params.~$\downarrow$ & GFLOPs~$\downarrow$ & Latency~$\downarrow$ & AP(\%)~$\uparrow$ \\
        \shline
        3-9-9-3                      & 53.40M & 86.28G & 2.60ms & 51.4 \\
        3-6-6-3                      & 50.92M & 79.70G & 2.11ms & 50.9 \\
        \ours{3-6-6-3 w/\small{CA}}  & \ours{52.30M} & \ours{79.90G} & \ours{2.40ms} & \ours{51.3} \\
        \end{tabular}}
    \end{minipage}
    
    ~\\

    \begin{minipage}[t]{.5\linewidth}
      \centering
      \subcaption{\small{Ablation study of backbone and neck proportions}}\label{tab:ablation:depth_width}
      \vspace{-3pt}
     \tablestyle{1pt}{1.2}
      \scalebox{0.95}{\begin{tabular}{ c | x{35}x{25}x{35}x{39}x{37}|x{35}}
        Model Size & Backbone & Neck & Params.~$\downarrow$ & GFLOPs~$\downarrow$ & Latency~$\downarrow$ & AP(\%)~$\uparrow$ \\
        \shline
        \ours{Small} & \ours{47\%} & \ours{45\%} & \ours{8.54M}  & \ours{15.76G} & \ours{\bd{1.21ms}} & \ours{\bd{43.9}} \\
        Small & 63\% & 29\% & 9.01M  & 15.85G & 1.37ms & 43.7 \\
        \shline
        \ours{Large} & \ours{47\%} & \ours{45\%} & \ours{50.92M} & \ours{79.70G} & \ours{\bd{2.11ms}} & \ours{50.9} \\
        Large & 63\% & 29\% & 57.43M & 93.73G & 2.57ms & \bd{51.0} \\
        \end{tabular}}
    \end{minipage}
    \begin{minipage}[t]{.48\linewidth}
        \centering
        \subcaption{\small{Design of the detection head}}\label{tab:ablation:head}
        \vspace{-3pt}
       \tablestyle{1pt}{1.2}
        \scalebox{0.95}{\begin{tabular}{ c | x{40}x{40}x{40}|x{40}}
          Head Type & Params.~$\downarrow$ & GFLOPs~$\downarrow$ & Latency~$\downarrow$ & AP(\%)~$\uparrow$ \\
          \shline
          Shared Head      & 52.32M & 80.23G & 2.44ms & 48.0\\
          Totally Separate & 57.03M & 80.23G & 2.44ms & 51.2\\
          \ours{Separate BN}      & \ours{52.32M} & \ours{80.23G} & \ours{2.44ms} & \ours{\bd{51.3}}\\
          \end{tabular}}
      \end{minipage}
  \end{table*}

\begin{table*}[!h]
  \centering
  \caption{\textbf{Ablation studies of label assignment} on COCO \texttt{val2017} set. The proposed setting is marked in gray
  }\label{table:ablation_label_assignment}
  \vspace{-6pt}
  \centering
  \begin{minipage}[t]{.32\linewidth}
    \centering
    \subcaption{\small{Ablation study of dynamic soft label assignment with ResNet50 1x schedule}}\label{tab:ablation:label_assignment}
    \vspace{-3pt}
      \tablestyle{1pt}{1.2}
       \scalebox{0.95}{\begin{tabular}{ x{35} x{35} x{35}|x{23}}
         Soft cls. cost & Soft ctr. prior& Log IoU cost & AP(\%)~$\uparrow$\\
         \shline
                        &                &               & 39.9 \\
         $\checkmark$   &                &               & 40.3 \\
         $\checkmark$   & $\checkmark$   &               & 40.8 \\
         \ours{$\checkmark$}   & \ours{$\checkmark$}   & \ours{$\checkmark$}  & \ours{\bd{41.3}} \\
         \end{tabular}}
    \end{minipage}\hspace{0.01\linewidth}
  \begin{minipage}[t]{.32\linewidth}
    \centering
    \subcaption{\small{Comparison with other label assignment with ResNet50 1x schedule}}\label{tab:ablation:compare_sota_label_assignment}
    \vspace{-3pt}
   \tablestyle{1pt}{1.2}
    \scalebox{0.95}{\begin{tabular}{ x{100} | x{33}}
      Method         & AP(\%)~$\uparrow$  \\
      \shline
      ATSS~\cite{ATSS}           & 39.2    \\
      PAA~\cite{PAA}            & 40.4    \\
      OTA~\cite{OTA}            & 40.7    \\
      TOOD~\cite{TOOD} (w/o T-Head) & 40.7    \\
      \ours{Ours}           & \ours{\bd{41.3}}   \\
      \end{tabular}}
  \end{minipage}\hspace{0.01\linewidth}
  \begin{minipage}[t]{.32\linewidth}
    \centering
      \subcaption{\small{Comparison with SimOTA label assignment on RTMDet-s with the same losses and other training strategies}}\label{tab:ablation:yolox_label_assignment}
    \vspace{-3pt}
   \tablestyle{1pt}{1.2}
    \scalebox{0.95}{\begin{tabular}{ x{100} | x{33}}
      Method  & AP(\%)~$\uparrow$ \\
      \shline
      SimOTA  & 43.2   \\
      \ours{Ours}    & \ours{\bd{44.5}}   \\
      \end{tabular}}
  \end{minipage}
\end{table*}

\subsection{Ablation Study of Model Arhitecture}
\paragraph{Large kernel matters.}
We first compare the effectiveness of different kernel sizes in the basic building block of CSPDarkNet~\cite{YOLOv4}, with kernel sizes ranging from 3$\times$3 to 7$\times$7.
A combination of 3$\times$3 convolution and 5$\times$5 kernel size depth-wise convolution achieves the optimal speed-accuracy trade-off (Table~\ref{tab:ablation:kernel_size}).

\paragraph{Balance of multiple feature scales.}
Using depth-wise convolution also increases the depth and reduces the inference speed.
Thus, we reduce the number of blocks in the 2nd and 3rd stages.
As revealed in Table~\ref{tab:ablation:num_blocks}, reducing the number of blocks from 9 to 6 results in a 20\% reduction of latency but decreases accuracy by 0.5\% AP.
To compensate for this loss in accuracy, we incorporate Channel Attention (CA) at the end of each stage, achieving a better speed-accuracy trade-off. 
Specifically, compared to the detector using 9 blocks in the second and third stages, the accuracy decreases by 0.1\% AP, but with a 7\% improvement in latency.
Overall, our modification successfully reduce the latency of the detector without sacrificing too much accuracy.

\paragraph{Balance of backbone and neck.}
Following \cite{YOLOv4, YOLOv5, YOLOv6, PPYOLOE},  we utilize the same basic block as the backbone for building the neck. We empirically study whether it is more economic to put more computations in the neck.
As shown in Table~\ref{tab:ablation:depth_width}, instead of increasing the complexity of the backbone, making the neck have similar capacity as the backbone can achieve faster speed with similar accuracy in both small and large real-time detectors.

\paragraph{Detection head.}
In Table~\ref{tab:ablation:head}, we compare different sharing strategies of the detection head for multi-scale features.
The results show that incorporating Batch Normalization (BN) into a shared-weight detection head causes a performance drop because of the statistical differences between different feature scales.
Using different detection heads for different feature scales can solve this issue but significantly increases the parameter numbers.
Using the same weights for different feature scales but different BN statistics yields the best parameter-accuracy trade-off.

\subsection{Ablation Study of Training Strategy}
\paragraph{Label assignment.}
We then verify the effectiveness of each component in the proposed dynamic soft label assignment strategy. Following previous conventions, we use SimOTA~\cite{YOLOX} as our baseline and employ the FocalLoss~\cite{lin2017_focal} and GIoU~\cite{giou}, which are the same as the training losses, as the cost matrix. 
As shown in Table~\ref{tab:ablation:label_assignment}, our baseline version can achieve an AP of 39.9\% on ResNet-50. 
Introducing IoU as a soft label in the classification cost improves the accuracy by 0.4\% AP, reaching 40.3\% AP. 
Replacing the fixed 3$\times$3 center prior with a softened center prior further improves the accuracy to 40.8\% AP.
By replacing the GIoU cost with the logarithm IoU cost, the model obtains 41.3\% AP.

The proposed label assignment strategy surpasses other high-performance strategies by 0.5\% AP on the same model architecture with the same losses (Table~\ref{tab:ablation:compare_sota_label_assignment}). 
When trained with a longer training schedule and stronger data augmentation, 
the proposed dynamic soft label assignment, together with the losses, surpasses SimOTA by 1.3\% AP on RTMDet-s (Table~\ref{tab:ablation:yolox_label_assignment}). 

\begin{table*}[t]
    \centering
    \caption{\textbf{Ablation studies of data augmentation} on COCO \texttt{val2017} set. The proposed settings are marked in gray
    }\label{table:ablation-data-aug}
    \centering
    \begin{minipage}[t]{.95\linewidth}
      \centering
      \subcaption{\small{Comparison with large-scale jittering (LSJ) and Mosaic \& MixUp of different setting. "small" means using a small cache size and FIFO popping method}}\label{tab:ablation:comparison-data-aug}
  \vspace{-3pt}
 \tablestyle{1pt}{1.2}
  \scalebox{0.95}{\begin{tabular}{ c | x{150} | x{150} |x{36}}
    Model & Data Aug. in 1st stage & Data Aug. in 2nd stage & AP(\%)~$\uparrow$ \\
    \shline
    \multirow{5}{*}{RTMDet-s} &  LSJ~\cite{scp}    &LSJ& 42.3   \\
     &  Mosaic \& MixUp  &  Mosaic \& MixUp                 & 41.9   \\
     &  Cached Mosaic \& MixUp  &  Cached Mosaic \& MixUp   & 41.9   \\
     &  Cached Mosaic \& MixUp  &LSJ     & 43.9   \\
     &  \ours{Cached(small) Mosaic \& MixUp} &\ours{LSJ} & \ours{\bd{44.2}}   \\
    \shline{}
    \multirow{5}{*}{RTMDet-l} &  LSJ &LSJ           & 46.7   \\
     &  Mosaic \& MixUp &  Mosaic \& MixUp               & 49.8   \\
     &  Cached Mosaic \& MixUp &  Cached Mosaic \& MixUp & 49.8   \\
     & \ours{Cached Mosaic \& MixUp} &\ours{LSJ}                 & \ours{\bd{51.3}}   \\
     &  Cached(small) Mosaic \& MixUp & LSJ & 51.1   \\
    \end{tabular}}
    \end{minipage}

    \begin{minipage}[t]{.48\linewidth}
      \centering
      \subcaption{\small{The speed comparison with vanilla and cached Mosaic \& MixUp}}\label{tab:ablation:comparison-cache}
        \vspace{-3pt}
        \tablestyle{1pt}{1.2}
        \scalebox{0.95}{\begin{tabular}{ x{50} | x{50} |x{55}}
         & Use cache & ms/100imgs~$\downarrow$ \\
        \shline
        Mosaic &                &  87.1  \\
        \ours{Mosaic} & \ours{$\checkmark$}   &  \ours{\bd{24.0}}  \\
        \shline
        MixUp  &                &  19.3  \\
        \ours{MixUp}  & \ours{$\checkmark$}   &  \ours{\bd{12.4}}  \\
    \end{tabular}}
    \end{minipage}  
    \begin{minipage}[t]{.48\linewidth}
      \centering
      \subcaption{\small{Comparison with data augmentation in YOLOX}}\label{tab:ablation:data-aug-yolox}
  \vspace{-3pt}
 \tablestyle{1pt}{1.2}
  \scalebox{0.95}{\begin{tabular}{ c | x{80} |x{33}}
    Model & Data Aug. & AP(\%)~$\uparrow$ \\
    \shline
    RTMDet-s &  YOLOX   & 42.9   \\
    \ours{RTMDet-s} &  \ours{Ours}    & \ours{\bd{44.2}}   \\
    \shline
    RTMDet-l &  YOLOX   & 50.6   \\
    \ours{RTMDet-l} &  \ours{Ours}    & \ours{\bd{51.3}}   \\
    \end{tabular}}
    \end{minipage}
  \end{table*}

\paragraph{Data augmentation.}
We then study different combinations of data augmentations at different training stages.
The first and second training stage takes 280 and 20 epochs, respectively. When the data augmentation is the same in these two stages, it essentially forms a one-stage training.
Following YOLOX, the range of random resizing in Mosaic for tiny and small models is $(0.5, 2.0)$, while $(0.1, 2.0)$ is used for larger models.
As shown in Table~\ref{tab:ablation:comparison-data-aug}, using large-scale jittering (LSJ)~\cite{scp} in all the stages is 0.4\% AP better than using MixUp and Mosaic. The effect of cached Mosaic and MixUp is consistent with the original ones when there are sufficient cached samples. Still, the cache mechanism speeds up Mosaic and MixUp by $\sim3.6\times$ and $\sim1.5\times$, respectively (Table~\ref{tab:ablation:comparison-cache}).

Using LSJ in the second stage instead of Mosaic and MixUp brings 2\% AP and 1.5\% improvement for RTMDet-s and RTMDet-l, respectively.
This indicates that Mosaic and MixUp is a stronger augmentation than LSJ but also introduces more noise in training, which should be thrown in the second stage.
We also observe that if the cache size was reduced to around 10 images and the First-In-First-Out (FIFO) popping method was applied, it is possible to mix the same image with different data augmentation operations in the same batch, which may have similar effects as repeated augmentation~\cite{berman2019multigrain} and can slightly improve tiny and small models (by approximately 0.5\% AP).

Compared with YOLOX, we avoid random rotation and shearing in the first training stage because they cause misalignment between box annotations and the inputs. %
Instead, we increase the number of mixed images from 5 to 8 in each training sample to keep the strength of data augmentation in the first stage.
Overall, the new combination of data augmentation explored in this paper is consistently better than those of YOLOX at different model sizes~\ref{tab:ablation:data-aug-yolox}.

\paragraph{Optimization strategy.}
We finally conduct experiments on the optimization strategies. 
The results in Table~\ref{tab:ablation:opt_strategy} indicate that SGD leads to unstable convergence progress with heavy data augmentation in training. As a result, we selected AdamW with a 0.05 weight decay and Cosine Annealing LR as our baseline.
To avoid overfitting in the early or middle training progress due to the quickly reduced learning rate by Cosine Annealing, we adapt a flat-cosine approach, where a fixed learning rate is used in the first half of training epochs, and cosine annealing is then used in the second half. This modification improves the performance by 0.3\% AP.
Furthermore, inhibiting weight decay on normalization layers and biases following previous practices~\cite{bag_of_tricks} brings 0.9\% AP improvement.
Finally, applying a pre-trained ImageNet backbone through the RSB~\cite{RossWightman2021ResNetSB} training strategy leads to a further 0.3\% increase in AP.
The above-mentioned tricks synergically yield significant improvements of 1.5\% AP.

\begin{table}[t]
  \centering
  \caption{\small{\textbf{Ablation study of optimization strategy} based on RTMDet-s. The proposed setting is marked in gray}}\label{tab:ablation:opt_strategy}
  \vspace{-3pt}
 \tablestyle{1pt}{1.2}
  \scalebox{0.95}{\begin{tabular}{ l |x{33}}
    Optimizer & AP(\%)~$\uparrow$ \\
    \shline
    SGD  + CosineLR                 & unstable   \\
    AdamW  + CosineLR               & 43.0   \\
    AdamW + Flat CosineLR           & 43.3   \\
    {\quad} + w/o norm\&bias decay  & 44.2   \\
    \ours{{\quad} + RSB pretrain}          & \ours{\bd{44.5}}   \\
    \end{tabular}}
\end{table}

\begin{table*}[h!]
  \centering
  \caption{\small{Step-by-step improvements from YOLOX-s baseline to RTMDet-s. The proposed setting is marked in gray}}\label{tab:ablation:step_by_step}
  \vspace{-3pt}
 \tablestyle{1pt}{1.2}
  \scalebox{0.95}{\begin{tabular}{ l | x{70}x{70}x{60} | x{40}}
    Model  & Params(M)~$\downarrow$ & FLOPs(G)~$\downarrow$ & Latency(ms)~$\downarrow$ & AP(\%)~$\uparrow$ \\
    \shline
    YOLOX baseline                             & 9.0M          & 13.4G       & 1.20ms  & 40.2        \\
    {\quad} + AdamW \& Flat CosineLR           & 9.0M          & 13.4G       & 1.20ms  & 40.6(+0.4)  \\
    {\quad} + New architecture                 & 10.07M(+1.07) & 14.8G(+1.4) & 1.22ms  & 41.8(+1.2)  \\
    {\quad} + SepBNHead                        & 8.89M(-1.18)  & 14.8G       & 1.22ms  & 41.8(+0.0)  \\
    {\quad} + Label Assign \& Loss             & 8.89M         & 14.8G       & 1.22ms  & 42.9(+1.1)  \\
    {\quad} + Imporved data augmentations      & 8.89M         & 14.8G       & 1.22ms  & 44.2(+1.3)  \\
    \ours{{\quad} + RSB pretrained backbone}          & \ours{8.89M}         & \ours{14.8G}      & \ours{1.22ms}  & \ours{\bd{44.5(+0.3)}}  \\
    \end{tabular}}
\end{table*}

\subsection{Step-by-step Results}
As demonstrated in Table~\ref{tab:ablation:step_by_step}, we have made successive modifications to YOLOX-s.
By modifying the optimization strategy, the model accuracy is improved by 0.4\%. 
The new architecture that has a similar capacity of backbone and neck, constructed by the new basic building block with large-kernel depth-wise convolutions, improves the model accuracy by 1.2\% AP with marginal latency costs.
Using a detection head with shared weights reduces the number of parameters significantly without hurting the accuracy. 
Subsequent enhancements to the label assignment strategy and training losses boost the performance by 1.1\% AP. The new combination of data augmentations and the pre-training of backbone leads to 1.3\% AP and 0.3\% AP  improvement, respectively.
The synergy of these modifications results in the RTMDet-s, which outperforms the baseline by 4.3\% AP.


\section{Conclusion}
In this paper, we empirically and comprehensively study each critical component in real-time object detectors, including model architectures, label assignment, data augmentations, and optimization. We further explore minimal adaptations of a high-precision real-time object detector for real-time instance segmentation and rotated object detection.
The findings in the study result in a new family \emph{\textbf{R}eal-\textbf{T}ime \textbf{M}odels for object \textbf{Det}ection}, named \textbf{RTMDet}, and its derivatives for different object recognition tasks.
RTMDet demonstrates a superior trade-off between accuracy and speed in industrial-grade applications, with different model sizes for different object recognition tasks. 
We hope RTMDet, with the experimental results, can pave the way for future research and industrial development of real-time object recognition tasks.

\appendix
\setcounter{table}{0} 
\setcounter{figure}{0}
\setcounter{equation}{0}
\renewcommand{\thetable}{A\arabic{table}}
\renewcommand\thefigure{A\arabic{figure}} 
\renewcommand\theequation{A\arabic{equation}}
\section{Appendix}\label{sec:Supplementary}

\subsection{Benchmark Results}

\begin{table*}[th]
  \centering
  \vspace{-3pt}
    \caption{\small{\textbf{Comparison of RTMDet-R with PPYOLOE-R} on the number of parameters, FLOPs, latency, and accuracy. The inference speeds of all models are measured in the same environment. The results of the proposed RTMDet-R is marked in gray. The best results are in bold}}\label{tab:rotate_speed}
 \tablestyle{1pt}{1.2}
  \scalebox{0.98}{\begin{tabular}{ c | x{60}x{50}x{50}x{55}|x{40}}
    Model       & Input shape & Params(M)~$\downarrow$ & FLOPs(G)~$\downarrow$  & Latency(ms)~$\downarrow$ & mAP(\%)~$\uparrow$\\
    \shline
    \ours{RTMDet-R-tiny} & \ours{1024}     & \ours{4.88}    & \ours{20.45}    & \ours{3.04}    & \ours{\textbf{75.36}} \\
    \shline
    PPYOLOE-R-s  & 1024     & 8.24   & 22.16   & 3.38       & 73.82            \\
    \ours{RTMDet-R-s}    & \ours{1024}     & \ours{8.86}   & \ours{37.62}   & \ours{3.44}    & \ours{\bd{76.93}} \\
    \shline
    PPYOLOE-R-m   & 1024     & 24.66   & 65.40     & 5.26       & 77.64           \\
    \ours{RTMDet-R-m}    & \ours{1024}     & \ours{24.67}  & \ours{99.76}  & \ours{5.79}    & \ours{\textbf{78.24}} \\
    \shline
    PPYOLOE-R-l   & 1024     & 55.17   & 145.88     & 7.09       & 78.14         \\
    \ours{RTMDet-R-l}    & \ours{1024}     & \ours{52.27}   & \ours{204.21}  & \ours{8.20}    & \ours{\bd{78.85}}           \\
     \shline
    PPYOLOE-R-x   & 1024     & 104.18   & 275.41     & 10.36       & 78.28       
    \end{tabular}}
\end{table*}

\begin{table*}[h]
\centering
  \vspace{-3pt}
    \caption{\small{\textbf{Comparison with state-of-the-art methods on DOTA v1.5 dataset.} MS means multi-scale training and testing. For the DOTA-v1.5 dataset, we use 4 NVIDIA A100 GPUs for training. Since we find that COCO pretraining significantly improves the results on DOTA-v1.5, we use COCO pretraining by default. DOTA-v1.5 has 16 different object categories: plane (PL), baseball diamond (BD), bridge (BR), ground track field (GTF), small vehicle (SV), large vehicle (LV), ship (SH), tennis court (TC), basketball court (BC), storage tank (ST), soccer ball field (SBF), roundabout (RA), harbor (HA), swimming pool (SP), helicopter (HC) and container crane (CC). The AP of each category is listed. The bold fonts indicate the best performance. The results of the proposed RTMDet-R are marked in gray}}\label{tab:dota1_5}
\resizebox{\textwidth}{!}{
\begin{tabular}{c|c|cccccccccccccccc|c}
Method & MS &  PL &  BD &  BR &  GTF &  SV &  LV &  SH &  TC &  BC &  ST &  SBF &  RA &  HA &  SP &  HC & CC & mAP\\ \hline
RetinaNet-OBB \cite{lin2017_focal} &  & 71.43 & 77.64 & 42.12 & 64.65 & 44.53 & 56.79 & 73.31 & 90.84 & 76.02 & 59.96 & 46.95 & 69.24 & 59.65 & 64.52 & 48.06 & 0.83 & 59.16  \\
FR-OBB \cite{ren2015faster} &  & 71.89 & 74.47 & 44.45 & 59.87 & 51.28 & 69.98 & 79.37 & 90.78 & 77.38 & 67.50 & 47.75 & 69.72 & 61.22 & 65.28 & 60.47 & 1.54 & 62.00  \\
MASK RCNN \cite{mask_rcnn} &  & 76.84 & 73.51 & 49.90 & 57.80 & 51.31 & 71.34 & 79.75 & 90.46 & 74.21 & 66.07 & 46.21 & 70.61 & 63.07 & 64.46 & 57.81 & 9.42 & 62.67  \\
HTC \cite{Chen_2019_CVPR} &  & 77.80 & 73.67 & 51.40 & 63.99 & 51.54 & 73.31 & 80.31 & 90.48 & 75.12 & 67.34 & 48.51 & 70.63 & 64.84 & 64.48 & 55.87 & 5.15 & 63.40  \\
DAFNe & $\checkmark$ & - & - & - & - & - & - & - & - & - & - & - & - & - & - & - & - & 71.99  \\
OWSR \cite{Li_2019_CVPR_Workshops} &  $\checkmark$ & - & - & - & - & - & - & - & - & - & - & - & - & - & - & - & - & 74.90  \\
ReDet &  & 79.20 & 82.81 & 51.92 & 71.41 & 52.38 & 75.73 & 80.92 & 90.83 & 75.81 & 68.64 & 49.29 & 72.03 & 73.36 & 70.55 & 63.33 & 11.53 & 66.86  \\
ReDet  &  $\checkmark$ & 88.51 & 86.45 & \bd{61.23} & 81.20 & \bd{67.60} & \bd{83.65} & \bd{90.00} & 90.86 & 84.30 & 75.33 & \bd{71.49} & 72.06 & 78.32 & 74.73 & 76.10 & 46.98 & 76.80  \\
\ours{RTMDet-R-tiny}  & \ours{} & \ours{77.79} & \ours{83.03} & \ours{48.45} & \ours{73.37} & \ours{59.33} & \ours{81.30} & \ours{88.89} & \ours{90.88} & \ours{80.73} & \ours{76.26} & \ours{51.81} & \ours{71.59} & \ours{75.81} & \ours{75.19} & \ours{54.36} & \ours{20.01} & \ours{69.30}  \\
\ours{RTMDet-R-tiny}  &  \ours{$\checkmark$} & \ours{88.14} & \ours{83.09} & \ours{51.80} & \ours{77.54} & \ours{65.99} & \ours{82.22} & \ours{89.81} & \ours{90.88} & \ours{80.54} & \ours{81.34} & \ours{64.64} & \ours{71.51} & \ours{77.13} & \ours{76.32} & \ours{72.11} & \ours{46.67} & \ours{74.98}  \\
\ours{RTMDet-R-s}  & \ours{} & \ours{80.05} & \ours{84.36} & \ours{50.65} & \ours{72.04} & \ours{59.54} & \ours{81.79} & \ours{89.22} & \ours{\bd{90.90}} & \ours{83.07} & \ours{76.27} & \ours{56.82} & \ours{72.13} & \ours{76.25} & \ours{77.04} & \ours{65.66} & \ours{32.84} & \ours{71.79}  \\
\ours{RTMDet-R-s}  &  \ours{$\checkmark$} & \ours{88.14} & \ours{85.82} & \ours{52.90} & \ours{82.09} & \ours{65.58} & \ours{81.83} & \ours{89.78} & \ours{90.82} & \ours{83.31} & \ours{82.47} & \ours{68.51} & \ours{70.93} & \ours{\bd{78.00}} & \ours{75.77} & \ours{73.09} & \ours{47.32} & \ours{76.02}  \\
\ours{RTMDet-R-m} & \ours{} & \ours{80.34} & \ours{86.00} & \ours{54.02} & \ours{72.98} & \ours{63.21} & \ours{82.09} & \ours{89.46} & \ours{90.87} & \ours{85.12} & \ours{76.69} & \ours{63.12} & \ours{72.14} & \ours{77.91} & \ours{76.04} & \ours{71.57} & \ours{32.24} & \ours{73.36}  \\
\ours{RTMDet-R-m} &  \ours{$\checkmark$} & \ours{89.07} & \ours{\bd{86.71}} & \ours{52.57} & \ours{82.47} & \ours{66.13} & \ours{82.55} & \ours{89.77} & \ours{90.88} & \ours{84.39} & \ours{83.34} & \ours{69.51} & \ours{73.03} & \ours{77.82} & \ours{75.98} & \ours{\bd{80.21}} & \ours{42.00} & \ours{76.65}  \\
\ours{RTMDet-R-l} & \ours{} & \ours{80.73} & \ours{84.79} & \ours{54.09} & \ours{76.30} & \ours{63.56} & \ours{83.06} & \ours{89.77} & \ours{90.89} & \ours{86.65} & \ours{76.98} & \ours{63.68} & \ours{70.31} & \ours{78.11} & \ours{75.91} & \ours{75.09} & \ours{31.20} & \ours{73.82}  \\
\ours{RTMDet-R-l} &  \ours{$\checkmark$} & \ours{\bd{89.31}} & \ours{86.38} & \ours{55.09} & \ours{\bd{83.17}} & \ours{66.11} & \ours{82.44} & \ours{89.85} & \ours{90.84} & \ours{\bd{86.95}} & \ours{\bd{83.76}} & \ours{68.35} & \ours{\bd{74.36}} & \ours{77.60} & \ours{\bd{77.39}} & \ours{77.87} & \ours{\bd{60.37}} & \ours{\bd{78.12}}  \\
\end{tabular}}
\end{table*}

\begin{table}[th]
  \centering
  \vspace{-3pt}
    \caption{\small{Comparison with state-of-the-art methods on HRSC2016 dataset. mAP$_{07}$ and mAP$_{12}$ indicate that the results were evaluated under VOC2007 and VOC2012 metrics (\%), respectively. We report both results for fair comparison. The results of the proposed RTMDet-R is marked in gray. The results of the proposed RTMDet is marked in gray. The best results are in bold}}\label{tab:hrsc}
 \tablestyle{1pt}{1.2}
  \scalebox{0.98}{\begin{tabular}{ c|x{50}|x{52}x{52}}
    Model   & Input shape & mAP$_{07}$ & mAP$_{12}$\\
    \shline
    RoI Trans. & 512-800 & 86.20 & - \\
    Gliding Vertex & 512-800 & 88.20 & - \\
    R$^{3}$Det & 800$\times$800 & 89.26 & 96.01 \\
    GWD & 800$\times$800 & 89.85 & 97.37 \\
    CSL & 800$\times$800 & 89.62 & 96.10 \\
    S$^{2}$ANet & 512-800 & 90.17 & 95.01 \\
    ReDet & 512-800 & 90.46 & \bd{97.63} \\
    Oriented RCNN & 800-1333 & 90.50 & 97.60 \\
    \ours{RTMDet-R-tiny}  & \ours{800$\times$800}     & \ours{\bd{90.60}}   & \ours{97.10}   \\
    \end{tabular}}
\end{table}

\paragraph{Comparison with PPYOLOE-R.} We further compare RTMDet-R with PPYOLOE-R in detail, and RTMDet-R is more competitive in accuracy and inference speed, as shown in Table~\ref{tab:rotate_speed}.
More surprisingly, RTMDet-R-m and RTMDet-R-l surpass PPYOLOE-R-l and PPYOLOE-R-x while being 18.5\% and 20.8\% faster, respectively. Code and models of RTMDet-R are released at MMRotate \cite{zhou2022mmrotate}.

\paragraph{Results on DOTA-v1.5.} We further verify the effectiveness of RTMDet-R on DOTA-v1.5 dataset. DOTA-v1.5 contains the same images as DOTA-v1.0 but annotates extremely small instances (less than 10 pixels) with 215k instances added, which makes it more challenging. For the DOTA-v1.5 dataset, we use 4 NVIDIA A100 GPUs for training. Since we find that COCO pretraining significantly improves the results on DOTA-v1.5, we use COCO pretraining by default. Other settings are consistent with DOTA-v1.0. As shown in Table~\ref{tab:dota1_5}, RTMDet-R-l surpasses the previous best method ReDet \cite{han2021redet} by 1.32\% mAP.

\paragraph{Results on HRSC2016.} We also verify RTMDet-R on HRSC2016 dataset, a ship detection dataset containing 1K images and a total of 2.9K ships collected from Google Earth. For the HRSC2016 dataset, we do not change the aspect ratios of images. We train all the models with 108 epochs for HRSC2016 dataset. Other settings are consistent with those for DOTA-v1.0. 
RTMDet-R also obtains a new state-of-the-art performance and achieves 90.6\% mAP$_{07}$ (Table~\ref{tab:hrsc}).


{\small
\bibliographystyle{ieee_fullname}
\bibliography{sections/mainbib}
}

\end{document}